# `ViconMAVLink`: A SOFTWARE TOOL FOR INDOOR POSITIONING USING A MOTION CAPTURE SYSTEM

Bo Liu[*]
Coordinated Science Laboratory University of Illinois at Urbana-Champaign Urbana, IL 61801
boliu1@illinois.edu

Normand Paquin
Coordinated Science Laboratory University of Illinois at Urbana-Champaign Urbana, IL 61801
paquin@illinois.edu

November 28, 2018

## ABSTRACT

Motion capture is a widely-used technology in robotics research thanks to its precise positional measurements with real-time performance. This paper presents `ViconMAVLink`, a cross-platform open-source software tool that provides indoor positioning services to networked robots. `ViconMAVLink` converts Vicon motion capture data into proper pose and motion data formats and send localization information to robots using the `MAVLink` protocol. The software is a convenient tool for mobile robotics researchers to conduct experiments in a controlled indoor environment.

Keywords  Motion capture · Vicon · Indoor positioning · Simulated GPS · UAS

## 1 Introduction

Motion capture is a technology for tracking and recording movement of objects [1]. High-performance motion capture systems, such as Vicon [2], have been widely-utilized in computer vision [3] as a validation tool and in mobile robotics research [4] as a localization method thanks to its fast frame rate and high precision.

A Vicon motion capture system tracks object based on markers. In a typical lab setup, three or more markers are attached in a unique geometric configuration to each rigid-body object for identification. Multiple low-latency cameras that are sensitive to the reflective rays from the markers are mounted to monitor a capture volume. A host computer is connected to all cameras processing reflective ray detections. The camera poses and absolute scale of the real world can be determined by calibration algorithms [5]. With calibrated cameras, the three-dimensional position of a marked object can be determined if the object is in view of multiple cameras at the same time. A typical Vicon system setup is capable of determining position with millimeter accuracy and with a measurement update frequency greater than 100 Hz.

`MAVLink` [6] is a light-weight messaging protocol widely-adopted by the open-source autopilot community [7]. A number of industrial-grade Unmanned Aerial System (UAS) systems, such as PX4 [8] and Ardupilot [9], use it as the underlying communication protocol. `MAVLink` implements useful pose-encoding messages for drones, such as `HIL_GPS`, which encodes hardware-in-the-loop GPS information, `LOCAL_POSITION_NED`, which encodes position and velocity information in the North-East-Down coordinate frame, and `ATT_POS_MOCAP`, which encodes attitude and position information used by motion capture systems.

The Intelligent Robotics Lab[1] at the University of Illinois, Urbana-Champaign provides a Vicon motion capture facility for robotics research. Vicon has provided proprietary programs such as Tracker [10] for simple motion capture recordings. However, they are primarily built to run on desktops with Intel processors. Vicon does not target their software on popular embedded computers such as the Raspberry Pi computer which uses ARM processors. Embedded computers are the driving force of recent mobile robotics research [11] due to their power efficiency and small form factor. To extend indoor positioning for mobile robots powered by embedded computers, the Vicon DataStream SDK [12] is utilized and `ViconMAVLink` was developed. `ViconMAVLink` is a cross-platform bridge software that provides indoor positioning measurements for robots. `ViconMAVLink` converts Vicon measurements into proper data formats defined by the `MAVLink` protocol. Cartesian coordinates and Simulated GPS localization results are computed and streamed over network to robots using the `MAVLink` protocol as the messaging layer. Thanks to the `MAVLink` communication layer, the robot may be agnostic about the origin of the localization data; as long as the data is a valid `MAVLink` localization packet, generated from real GPS signal or simulated data from `MAVLink`, the robot can utilize it. Mobile robotics researchers may utilize off-the-shelf robot platforms that already use `MAVLink`. Fortunately, the open-source community has provided abundant selections of such systems for the benefit of the research community.

---
[*]Now at NVIDIA Corp.
[1]robotics.illinois.edu

## 2 Method

In this section, we describe the methods underlying `ViconMAVLink`. To be concise, we discuss key mechanisms and techniques but may omit implementation details. The reader may study the source code of `ViconMAVLink`[13] for further details.

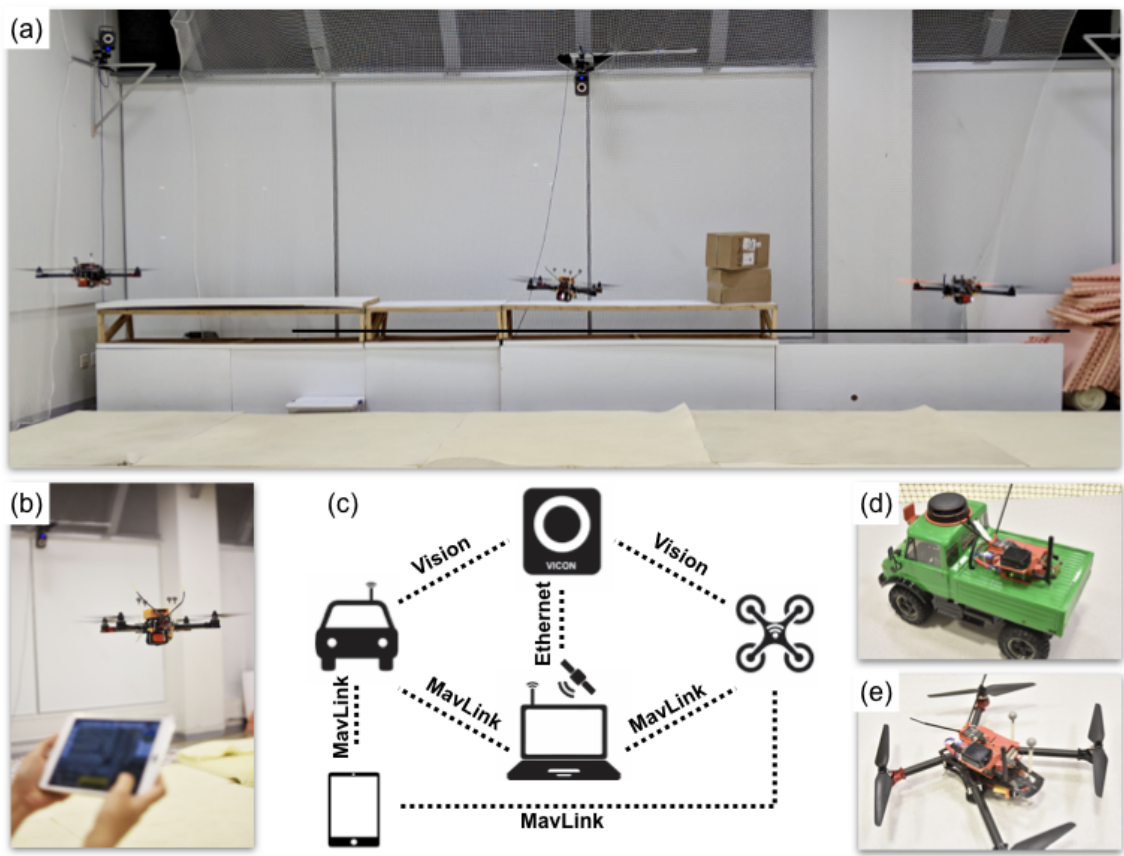

Figure 1: Lab setup. (a) An indoor capture volume provides a controlled environment for UAS experiments. Networked Vicon cameras are configured to monitor the capture volume. The floor is covered with cushioning pads to protect hardware assets. Safety netting is used to isolate people from the flying space. (b) The `MAVLink` messaging layer is widely-used by many open-source ground-control applications, such as the cross-platform `qgroundcontrol` program [14]. Users may monitor and operate `MAVLink`-enabled robots with `qgroundcontrol` running on a tablet. (c) Vicon cameras are connected via ethernet cables to a host desktop which fetches motion capture measurements from the cameras. `ViconMAVLink` runs on the same desktop for motion tracking using a Kalman filter [15]. Motion tracking results packed in proper `MAVLink` formats are sent to robots via Wi-Fi connections. (d) and (e) are Raspberry Pi-powered ground and aerial robots, respectively. They run the PX4 open-source autopilot software, receiving indoor positioning information from `ViconMAVLink`.

### 2.1 Network and Communication

As shown in figure 1 (c), Vicon cameras are connected to a host computer via ethernet cables. The host computer runs the Tracker [10] program to process data from the cameras. On the same host computer, `ViconMAVLink` periodically extracts Vicon results from Tracker via a User Datagram Protocol (UDP) loopback connection. `ViconMAVLink` computes simulated GPS, local position and motion capture results and converts them into proper `MAVLink` packets, which are sent to robots via wireless UDP connections. The robots may run different autopilot software but they must be able to parse `MAVLink` packets via a UDP port. For example, a Linux robot running the PX4 [8] autopilot may open a UDP port to listen to `MAVLink` packets. UDP communication has low-latency and only requires the sender to know the address of the receiver but not vice versa. This means that `ViconMAVLink` must know the address and port of the robot but the robot does not need to know the origin of the positioning data, i.e. the robot is agnostic to how the data is generated. This behavior allows minimum modifications to the robot and therefore provides convenience for mobile robotics experiments. The `MAVLink` messaging layer is also used for communication between robots and human operators. Because the `MAVLink` protocol already supports control messages, we may monitor and control the robots using a networked device. The only requirement is that the control application can parse `MAVLink` messages. For example, the `qgroundcontrol` program [14] can run on a tablet and is used to control a quadcopter, shown in figure 1 (b).

## 2.2 Software Architecture and Implementations

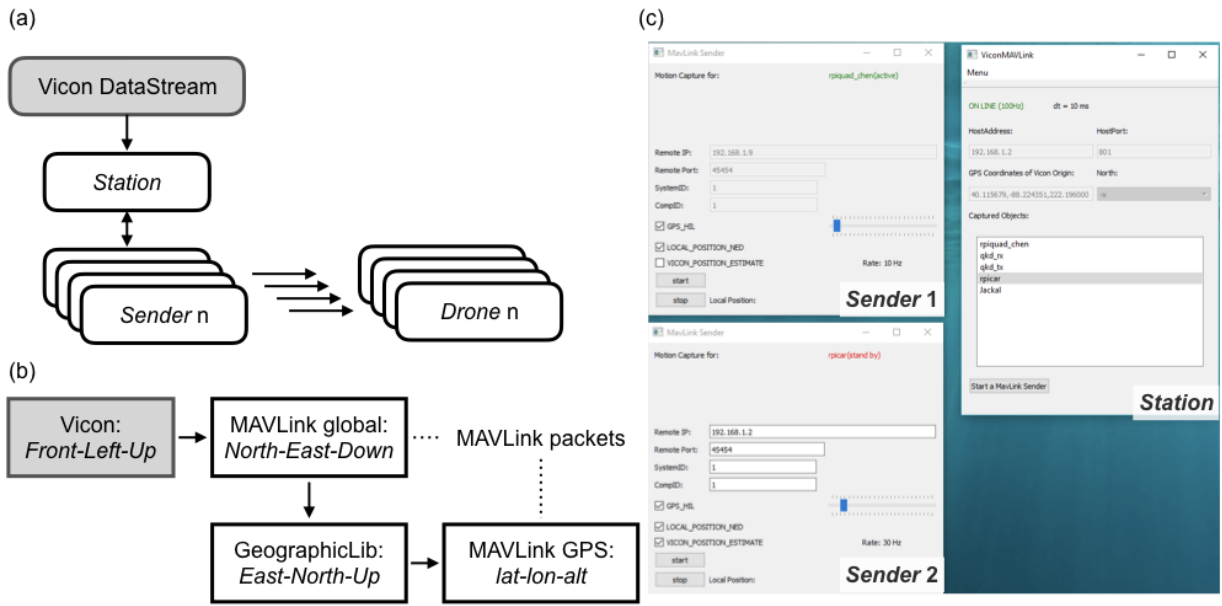

Figure 2: `ViconMAVLink` mechanisms. (a) Data flow of `ViconMAVLink`. The Station object fetches Vicon measure- ments and communicates with multiple Sender objects. Each Sender object is responsible for sending positioning data to respective drones. (b) Coordinate transforms in `ViconMAVLink`. (c) Screenshots of a running `ViconMAVLink`.

At runtime, there are three classes of interacting objects. First, `ViconMAVLink` starts a Station object which handles global tasks such as communication with Vicon, initialization of windows and parameters. Second, the Station object may launch multiple Sender objects, each of which communicates with the Station to get capture information about a specific object. To synchronize data, Qt read-write locking is used for data sharing between the Station (the writer) and Sender objects (the readers). Finally, we use Drone objects to encapsulate information about corresponding robots. Figure 2 (a) summarizes the data flow in `ViconMAVLink`.

`MAVLink` message types `HIL_GPS` and `LOCAL_POSITION_NED` have velocity fields. However, native Vicon measurements only provide per-frame position and orientation information. To recover the latent velocity information, we use a Kalman Filter [15] in the Sender class to track the robot's motion. Based on our tests (see Section 3) a generic linear Kalman Filter performs well enough for positioning indoor robots; a future improvement to `ViconMAVLink` is to implement model-specific motion tracking method for different robots.

We use `GeographicLib` [16] to convert Cartesian coordinates to GPS coordinates. `GeographicLib` uses East-North-Up axis mapping. Unfortunately, `MAVLink` and Vicon respectively use different axis mappings. `ViconMAVLink`inter- nally does axis mapping conversions for the three. First, the robot's position in Vicon coordinate frame is transformed into a proper format in `MAVLink` coordinate frame. To determine simulated GPS coordinates, we first determine `GeographicLib` local position coordinates in a Cartesian coordinate frame that is then used to compute the GPS coordinates using `GeographicLib`functions. Finally, we package data into proper `MAVLink`packets and send them to robots. The flow of coordinate transformations is summarized in figure 2 (b).

Figure 2 (c) shows sample screenshots of a running `ViconMAVLink`. The Station window displays global capture information from Vicon, such as the current update frequency and a list of robots that are available for tracking. The user may launch a Sender window for each robot. The frequency of a Sender sending positioning packets to a robot can be adjusted on the Sender window, which is independent of the global Vicon update frequency. There are check boxes to select which types of positioning packets to send to the robot. We use Qt signals and slots mechanism [17] for communication among interacting objects.

## 3 Evaluation

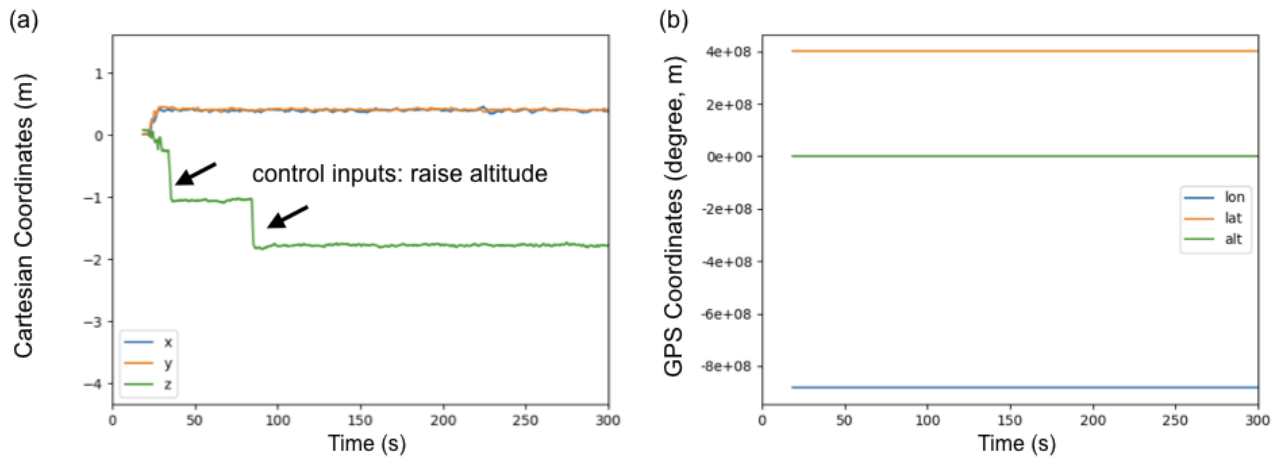

Figure 3: Evaluation of indoor positioning using simulated GPS. (a) A three-minute flight recording in Cartesian coordinates of a `MAVLink`-enabled UAS demonstrated a steady hovering performance. During the flight, the UAS is flying under a position-hold mode. i.e. without control input the UAS will hold its 3D position. The only localization data available to the UAS is the simulated GPS data sent by `ViconMAVLink`. The UAS operator gave two control inputs, highlighted by black arrows, to change the altitude of the UAS. The heights are represented by negative numbers due to the North-East-Down axis mapping in the `MAVLink` coordinate frame. (b) The same flight recording in GPS coordinates.

The `ViconMAVLink` software has been tested at the Intelligent Robotics Lab, University of Illinois at Urbana-Champaign using a $10 \times 10 \times 4$ m$^3$ indoor motion capture arena with 8 Vicon T-40 cameras running Tracker version 3.6 and DataStream SDK version 1.7. `ViconMAVLink` were tested on a Windows 10 PC with Vicon measurements fetched via Ethernet and a Ubuntu 16.04 laptop with Vicon measurements fetched via 2.4GHz Wi-Fi. A range of 100 - 240Hz camera update frequencies were tested to provide positioning data for 1 - 5 mobile robots powered by Raspberry Pi3 embedded computers. In all cases, the frame-drop rates are below 1%. In addition, our extensive research experiments using `ViconMAVLink` for autonomous quadcopters flying indoors have validated the usability of the software. Figure 3 shows one demo flight recording of a quadcopter hovering under a position-hold mode inside a lab where GPS signal is denied. The only localization measurements received by the quadcopter is the simulated GPS data streamed by `ViconMAVLink`. The standard deviations of position hold along x, y and z axes are 1.64, 1.15 and 1.27 cm, respectively. This centimeter position hold performance has shown the usefulness of `ViconMAVLink` for indoor positioning of mobile robots.

## 4 Acknowledgement

The authors acknowledge the support of the Coordinated Science Laboratory, the Departments of Aerospace Engineering, Computer Science, Electrical and Computer Engineering, and the Mechanical Science and Engineering and the College of Engineering at the University of Illinois at Urbana-Champaign for their contributions in support of this work. The work was carried out at the Intelligent Robotics Laboratory, Coordinated Science Laboratory, University of Illinois at Urbana-Champaign.